\documentclass{article}

\usepackage{microtype}
\usepackage{graphicx}
\usepackage{subcaption}
\usepackage{multirow}
\usepackage{colortbl}
\usepackage{pifont}
\usepackage{booktabs}
\usepackage[most]{tcolorbox}
\usepackage{listings}
\lstdefinestyle{promptstyle}{
  basicstyle=\ttfamily\small,
  breaklines=true,
  columns=fullflexible,
  keepspaces=true,
  showstringspaces=false
}
\usepackage{hyperref}

\usepackage[preprint]{icml2026}

\usepackage{amsmath}
\usepackage{amssymb}
\usepackage{amsfonts}
\usepackage{mathtools}
\usepackage{amsthm}
\usepackage{enumitem}

\usepackage[capitalize,noabbrev]{cleveref}

\theoremstyle{plain}

\theoremstyle{definition}

\theoremstyle{remark}

\usepackage[textsize=tiny]{todonotes}

\icmltitlerunning{\textit{ResGo} \& \textit{SurGo-R1}}

\begin{document}

\twocolumn[
  \icmltitle{\textit{SurGo-R1}: Benchmarking and Modeling \\Contextual Reasoning for Operative Zone in Surgical Video}



  \icmlsetsymbol{equal}{*}

  \begin{icmlauthorlist}
    \icmlauthor{Guanyi Qin}{equal,nus,grtii}
    \icmlauthor{Xiaozhen Wang}{equal,smu}
    \icmlauthor{Zhu Zhuo}{equal,nus}
    \icmlauthor{Chang Han Low}{nus}\\
    \icmlauthor{Yuancan Xiao}{smu}
    \icmlauthor{Yibing Fu}{nus}
    \icmlauthor{Haofeng Liu}{nus}
    \icmlauthor{Kai Wang}{smu}
    \icmlauthor{Chunjiang Li}{smu}
    \icmlauthor{Yueming Jin}{nus}
  \end{icmlauthorlist}

  \icmlaffiliation{nus}{National University of Singapore, Singapore}
  \icmlaffiliation{smu}{Southern Medical University, China}
  \icmlaffiliation{grtii}{Guangzhou Research Translation and Innovation Institute, National University of Singapore, China}

  \icmlcorrespondingauthor{Yueming Jin}{ymjin@nus.edu.sg}

  \icmlkeywords{Machine Learning, ICML}

  \vskip 0.3in
]



\printAffiliationsAndNotice{}  

\begin{abstract}
Minimally invasive surgery has dramatically improved patient operative outcomes, yet identifying safe operative zones remains challenging in critical phases, requiring surgeons to integrate visual cues, procedural phase, and anatomical context under high cognitive load. Existing AI systems offer binary safety verification or static detection, ignoring the phase-dependent nature of intraoperative reasoning. We introduce \textbf{\textit{ResGo}}, a benchmark of laparoscopic frames annotated with Go Zone bounding boxes and clinician-authored rationales covering phase, exposure quality reasoning, next action and risk reminder. We introduce evaluation metrics that treat correct grounding under incorrect phase as failures, revealing that most vision-language models cannot handle such tasks and perform poorly. We then present \textbf{\textit{SurGo-R1}}, a model optimized via RLHF with a multi-turn \textbf{phase-then-go} architecture where the model first identifies the surgical phase, then generates reasoning and Go Zone coordinates conditioned on that context. On unseen procedures, SurGo-R1 achieves 76.6\% phase accuracy, 32.7 mIoU, and 54.8\% hardcore accuracy, a 6.6$\times$ improvement over the mainstream generalist VLMs.
Code, model and benchmark will be available at \url{https://github.com/jinlab-imvr/SurGo-R1}.
\end{abstract}

\section{Introduction}

\begin{figure*}[!ht]
    \centering
    \includegraphics[width=\textwidth]{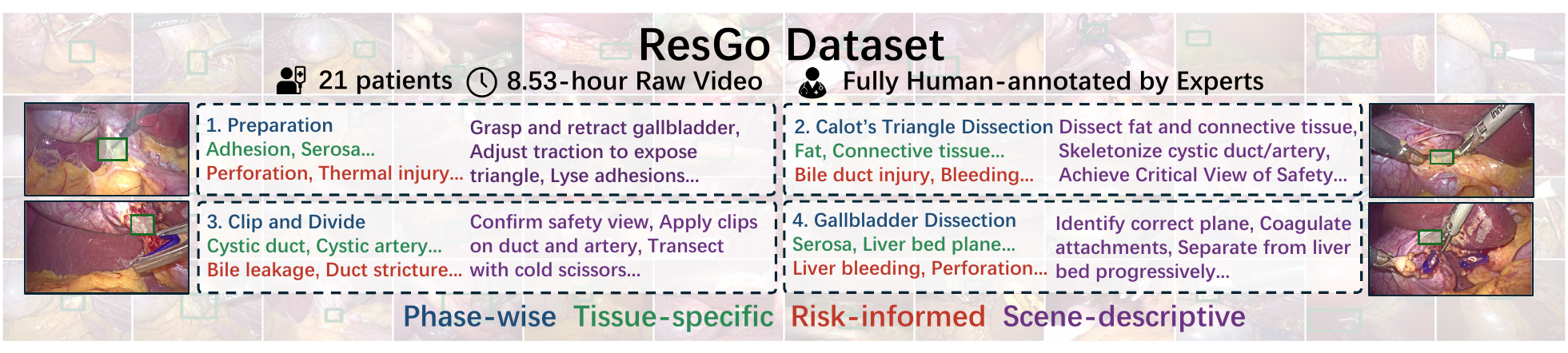}
    \caption{Demo of the proposed \textit{ResGo} dataset, a novel multimodal benchmark that covers phase recognition, Go Zone grounding, and safety reasoning. Built on \textit{ResGo}, a GRPO-optimized VLM with context-aware multi-turn reasoning, named \textit{SurGo-R1}, is further introduced, enabling safety-aware reasoning and Go Zone grounding across scenarios.}
    \label{fig:demo}
\end{figure*}

Minimally invasive surgery (MIS) has demonstrably improved patient outcomes and care quality, offering patients reduced postoperative pain, shorter hospital stays, and faster recovery compared to open surgery approaches \citep{velanovich2000laparoscopic}. 
Although enhanced visualization through high-resolution imaging is available, the inherent complexity of human anatomy still increases the demands on surgeons, imposing a heavier intraoperative cognitive load and requiring substantially more training and practice
\citep{way2003causes, zheng2010measuring, zegers2011incidence, magrina2002complications}. 
As one of the most crucial MIS procedures, cholecystectomy, for instance, still imposes significant cognitive demands despite its relatively standardized workflow. Performed over two million times annually in the U.S. and Europe alone \citep{pucher2018outcome, peery2022burden}, it carries a persistent risk of bile duct injury (BDI) which might result in a 2.5-fold increase in long-term mortality and often requires multiple reconstructive surgeries \citep{brunt2020safe}. Critically, most of these injuries stem from visual misperception \citep{way2003causes, strasberg2010rationale}: surgeons transect what they believe to be the cystic duct, only to discover that they have injured the common bile duct. This nature of struggling to interpret when anatomical landmarks are obscured by inflammation or aberrant anatomy, especially under heavy cognitive load, highlights the need for training and intraoperative guidance. 

Several core concepts have been proposed to ensure surgical safety in the community.
The most widely adopted strategy is to require surgeons to fulfill anatomical criteria before operation, such as Critical View of Safety (CVS) for cholecystectomy \citep{brunt2020safe}.
While this has shown promise, its effectiveness depends on subjective visual assessment under challenging conditions.
Recently, the concept of safety zones emerged, prioritizing a more direct identification of current safe operating regions \citep{madani2022artificial}. 
This highlights the need to develop systems that can provide context-aware support to surgeons.
Artificial Intelligence (AI) is increasingly transforming surgical care, and the emergence of AI copilots is opening up new possibilities. Conceived as cognitive collaborators for surgical education and intraoperative assistance, such systems can augment surgeons’ expertise for higher quality.

Early works primarily focused on binary safety verification, such as automated assessment of the CVS \citep{mascagni2022artificial} and subsequent graph-based extensions \citep{murali2023latent}. These approaches demonstrated the feasibility of AI-assisted safety evaluation but offered limited insight into \textit{where} safe dissection could occur. Building on this foundation, researchers \citep{madani2022artificial} pioneered the explicit visual grounding of safe zones, reframing surgical safety as a spatial understanding problem. By localizing safe dissection regions within the operative field, this work marked a key transition from binary verification to visual task formulation. Subsequent studies validated this paradigm against expert consensus \citep{laplante2023validation} and demonstrated its relevance in real bile duct injury cases \citep{khalid2023use}. Although spatial grounding marks a major advance over binary safety checks, existing methods exhibit limited generalization beyond the specific phase of dissection, and primarily produce basic visual recognition results, lacking the proactive, explainable capabilities to effectively copilot with surgeons.

To bridge this gap, we propose to leverage the ability of Vision-Language Models (VLMs) to connect natural language with complex visual scenes \citep{liu2023visual} and reason safety zones, which can augment human decision-making by advancing from \textit{simple visual recognition} to \textit{proactive and explainable prediction} \citep{li2023llava, jin2024surgical, zeng2025surgvlm}. 
We, as the first, introduce \textbf{\textit{ResGo}}, a multimodal dataset for cholecystectomy curated in-the-wild, that provides explainable Go Zone annotations to support understanding. The dataset is organized around clinically meaningful operative moments and offers hierarchically structured supervision. 
Specifically, each sample is spatially grounded with Go Zone bounding boxes. Moreover, this spatial evidence is \textit{\textbf{paired}} with clinician-authored, interpretable annotations, including the ongoing surgical phase, a concise description of the current operative action, key safety points that emphasize anatomical risks and disallowed maneuvers, and an explicit next-step statement.
This design enables models to learn not only \textit{where} the Go Zone lies, but also \textit{why} it is considered safe and \textit{how} the operative plan should proceed, facilitating transparency of safety-aware guidance. 
The rich nature of information of \textit{ResGo} enables comprehensive support for modern VLMs development, accommodating diverse training and evaluation patterns including supervised fine-tuning and RLHF, furthering agentic workflows, while remaining faithfully aligned with real-world practice and surgeon experience.

Building on \textit{ResGo}, a baseline named \textbf{\textit{SurGo-R1}} is also proposed. Given an intraoperative image, SurGo-R1 is trained with GRPO to generate structured, clinically grounded outputs that operationalize Go Zone understanding beyond pixel classification. 
Concretely, the model reasons about four complementary components: (1) \textit{Location}, a textual localization that describes where the Go Zone lies relative to salient anatomical landmarks; (2) \textit{Exposure}, an assessment of whether current retraction and dissection have achieved sufficient visualization for safe progression; (3) \textit{Next Action}, the immediate recommended maneuver consistent with safe advancement toward CVS; and (4) \textit{Critical Risk}, the primary source of potential misperception or injury risk in the current context. Across held-out procedures, SurGo-R1 achieves strong performance in both spatial localization and structured reasoning tasks. In summary, our primary contributions are as follows:
\begin{itemize}[itemsep=2pt, parsep=2pt, topsep=2pt, partopsep=2pt, leftmargin=1.5em]
    \item We introduce \textbf{\textit{ResGo}}, the first benchmark for cholecystectomy that pairs Go Zone localization with clinician-authored rationales for phase-dependent surgical reasoning annotations, aligning perception for safety-aware learning.
    \item Upon \textit{ResGo}, a novel practice-aligned contextual reasoning pattern as \textbf{phase-then-go} is proposed, accompanied by corresponding evaluation protocols, promoting agentic or complex reasoning research.
    \item We propose \textbf{\textit{SurGo-R1}}, a reasoning-based VLM optimized using GRPO that goes beyond static detection by producing structured, interpretable guidance, reflecting how surgeons assess and decide safe actions.
    \item \textit{SurGo-R1} substantially improves generalization on held-out procedures, achieving 76.6\% phase accuracy and 32.7 mIoU, and outperforming generalist VLM baselines with 6.6$\times$.
\end{itemize}

\section{Related Work}

\textbf{Surgical Safety Intelligence}
Bile duct injury during laparoscopic cholecystectomy predominantly results from visual misperception rather than technical error \citep{way2003causes}, motivating computational approaches to augment surgeon awareness. The CVS framework provides anatomical criteria for safe clip application, with DeepCVS \citep{mascagni2022artificial} pioneering automated verification and LG-CVS \citep{murali2023latent} enhancing robustness through graph-based anatomical reasoning. The Endoscapes dataset \citep{mascagni2025endoscapes} further advanced the field with 201 densely annotated cholecystectomy videos. Complementing CVS assessment, \citet{madani2022artificial} introduced the zone-based paradigm, reframing surgical safety as spatial localization of safe versus dangerous dissection zones, with subsequent validation against expert panels \citep{laplante2023validation} and on actual injury cases \citep{khalid2023use}. However, these methods operate as fixed-output systems requiring explicit labels for each target category, unable to interpret implicit safety queries or provide knowledge-grounded explanations.

\textbf{Reasoning Grounding}
Vision-language models have enabled sophisticated connections between natural language and visual content \citep{liu2023visual, dai2023instructblip}. Medical adaptations \citep{li2023llava, jin2024surgical, zeng2025surgvlm} show promise on clinical visual QA but generate purely textual outputs without spatial grounding. Visual grounding approaches have evolved from phrase-based localization \citep{plummer2015flickr30k} to VLM-integrated methods producing bounding boxes \citep{peng2023kosmos, chen2023shikra} or region-level understanding \citep{you2023ferret, rasheed2024glamm}. The reasoning grounding paradigm \citep{lai2024lisa} advances further by handling implicit queries requiring world knowledge, with extensions enhancing multi-target capabilities \citep{yang2023lisa++, ren2024pixellm}. Emerging medical reasoning grounding efforts \citep{huang2025medseg, tong2025medisee, liu2025gemex} target diagnostic imaging rather than procedural guidance. Our work introduces reasoning grounding to intraoperative surgery, where the model localizes safe dissection regions while generating interpretable responses.

\section{\textit{ResGo} Benchmark}
To advance surgical reasoning capabilities, we introduce \textit{ResGo} dataset. Unlike existing in-domain benchmarks limited to basic perception tasks such as phase recognition \citep{nwoye2023cholectriplet2021,twinanda2016endonet}, \textit{ResGo} annotates the underlying rationale behind safe dissection and Go Zone identification. By capturing this cognitive process alongside visual grounding, the dataset supports both intraoperative guidance and surgical education, demonstrating not only what is seen but how safe surgery is performed.

\subsection{Source Acquisition and Demographics}
To ensure clinical grounding and real-world in-vivo variability, we curated 21 laparoscopic cholecystectomy videos from collaborating institutions. Procedures were recorded with informed consent and ethical approval for education, training, and research purposes using 4K imaging systems (Olympus, Storz, and Huanuokang) to preserve fine-grained anatomical details. In accordance with the \textit{Declaration of Helsinki}, all videos were de-identified to protect patient privacy\footnote{Ethical approval details will be provided upon publication.}. Raw footage was standardized to $1920 \times 1080$ resolution and sampled at 0.2 fps, yielding 6,138 frames.

\begin{figure*}[!t]
    \centering
    \includegraphics[width=\linewidth]{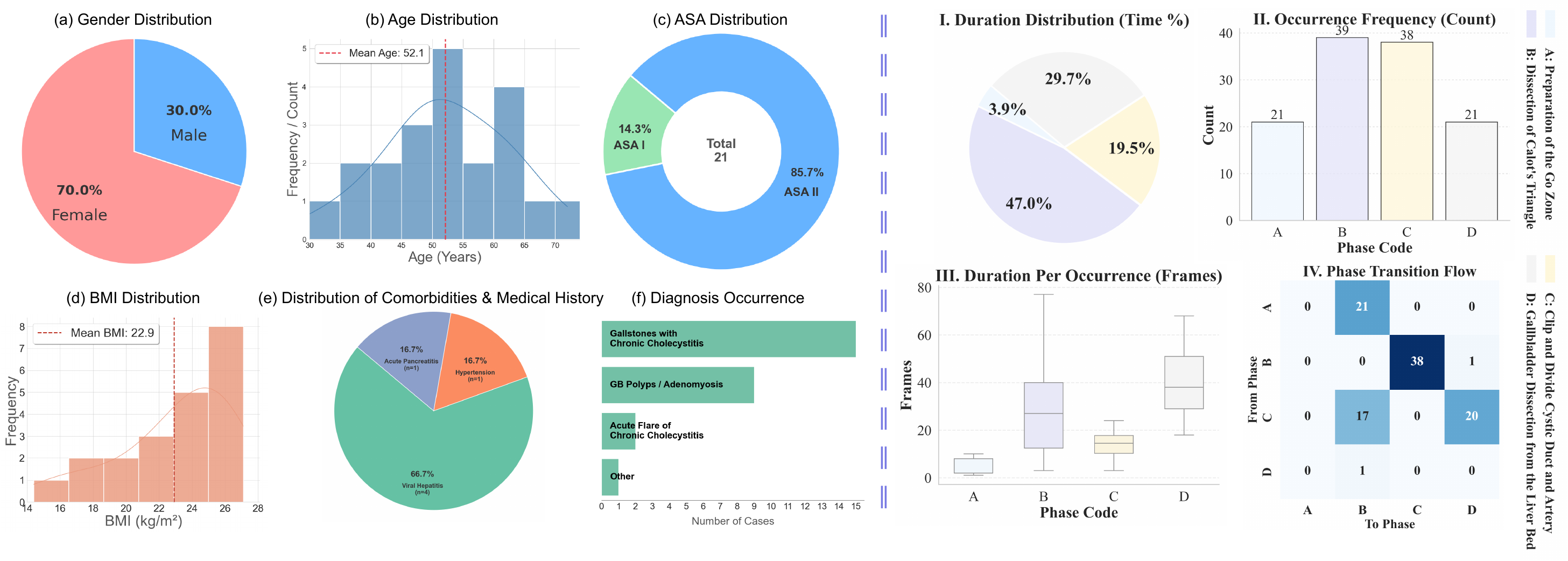}
    \caption{The left panels detail patient demographics and clinical profiles, including distributions for gender, age, ASA classification, BMI, and relevant medical history. The right panels present the processed surgical phase statistics, illustrating duration distributions, occurrence frequencies, and the phase transition matrix for the four defined phases.}
    \label{fig:all_stats}
\end{figure*}

Notably, the dataset encompasses diverse patient demographics, as shown in Fig. \ref{fig:all_stats}. Patients are predominantly female (70\%), with ages ranging from 30 to over 70 years (peak $\sim$55 years) and BMI from 14 to 27, consistent with typical in-the-wild cholecystectomy populations. The cohort also spans varied clinical presentations: 85\% were ASA Class 2 (mild systemic disease), with 6 of 21 patients presenting pre-existing conditions including hypertension, diabetes, and viral hepatitis (4 cases). One patient presented with concurrent acute pancreatitis. Clinical diagnoses such as gallstones with chronic cholecystitis were also recorded. By incorporating such a diverse array of patient demographics, comorbidities, and pathological variations, the dataset is positioned to expose models to the heterogeneity inherent in routine clinical practice, thereby establishing a robust foundation for training VLMs capable of adapting to the unpredictable nature of live surgery.

\begin{table}[!t]
    \centering
    \caption{\textbf{Annotation Statistics and Properties.} 2,686 annotated samples in total, where each sample consists of a structured instruction spanning four dimensions. \textbf{desc.} for description.}
    \label{tab:dataset_stats}
    \resizebox{\columnwidth}{!}{
    \begin{tabular}{l l}
        \toprule
        \textbf{Attribute} & \textbf{Specification} \\
        \midrule
        \multicolumn{2}{l}{\textit{\textbf{Data Acquisition}}} \\
        Equipment & Laparoscope (Olympus/Storz/H.) \\
        Video Resolution & $1920 \times 1080$ (Downsampled) \\
        FPS & 0.2 fps \\
        Total Videos & 21 (Cholecystectomy) \\
        Total Raw Frames & 6,138 \\
        Avg. Duration & $\sim$ 25 mins / video \\
        \midrule
        \multicolumn{2}{l}{\textit{\textbf{Annotation Scope}}} \\
        Annotators & 3 Senior Surgeons ($>$3 years exp.) \\
        Reviewers & 3 Chief Surgeons ($>$15 years exp.) \\
        \textbf{Total Annotated} & \textbf{2,686 Samples} \\
        \midrule
        \multicolumn{2}{l}{\textit{\textbf{Annotation Details (4 Dimensions per Sample)}}} \\
        a. Surgical Phase & Current procedural phase \\
        b. Anatomic & (1) Text: Grounding desc. of Go Zones \\
                            & (2) Visual: \texttt{LabelMe} Bounding box \\
        c. Reasoning & Analysis upon quality of exposure \\
        d. Planning & Next-step \& critical risk reminders \\
        \bottomrule
    \end{tabular}
    }
\end{table}

\subsection{Annotation Framework and Pipeline}

To ensure clinical relevance and precision, a rigorous annotation pipeline was implemented by six recruited hepatobiliary surgeons. Prior to the commencement of detailed labeling, a preliminary screening was performed to filter out frames deemed operationally irrelevant, and, to ensure procedural consistency, the temporal boundaries of each video clip were standardized to commence with the preparation of Go Zone exposure and conclude with the gallbladder dissection from the liver bed. Consequently, a curated set of 2,686 high-quality frames was yielded for subsequent annotation.

Before the annotation process was initiated, a consensus meeting was convened among the experts to establish unified annotation protocols. Particular emphasis was placed on standardizing the criteria for the Go Zone, which, after consensus and agreement, defined as the \textbf{\textit{optimal region for surgical operation within the current field of view in strict adherence to safety guidelines}}, while, for the preparation or exposure stage, the Go Zone is defined as the targeted anatomical region requiring clearance to facilitate the safe identification. Subsequently, the workload was distributed by video to three senior surgeons, each possessing over three years of experience. Following this allocation, the annotation was conducted across four distinct dimensions in accordance with the established pipeline, as illustrated in Fig.~\ref{fig:anpipe} as:
\begin{figure}[!t]
    \centering
    \includegraphics[width=\linewidth]{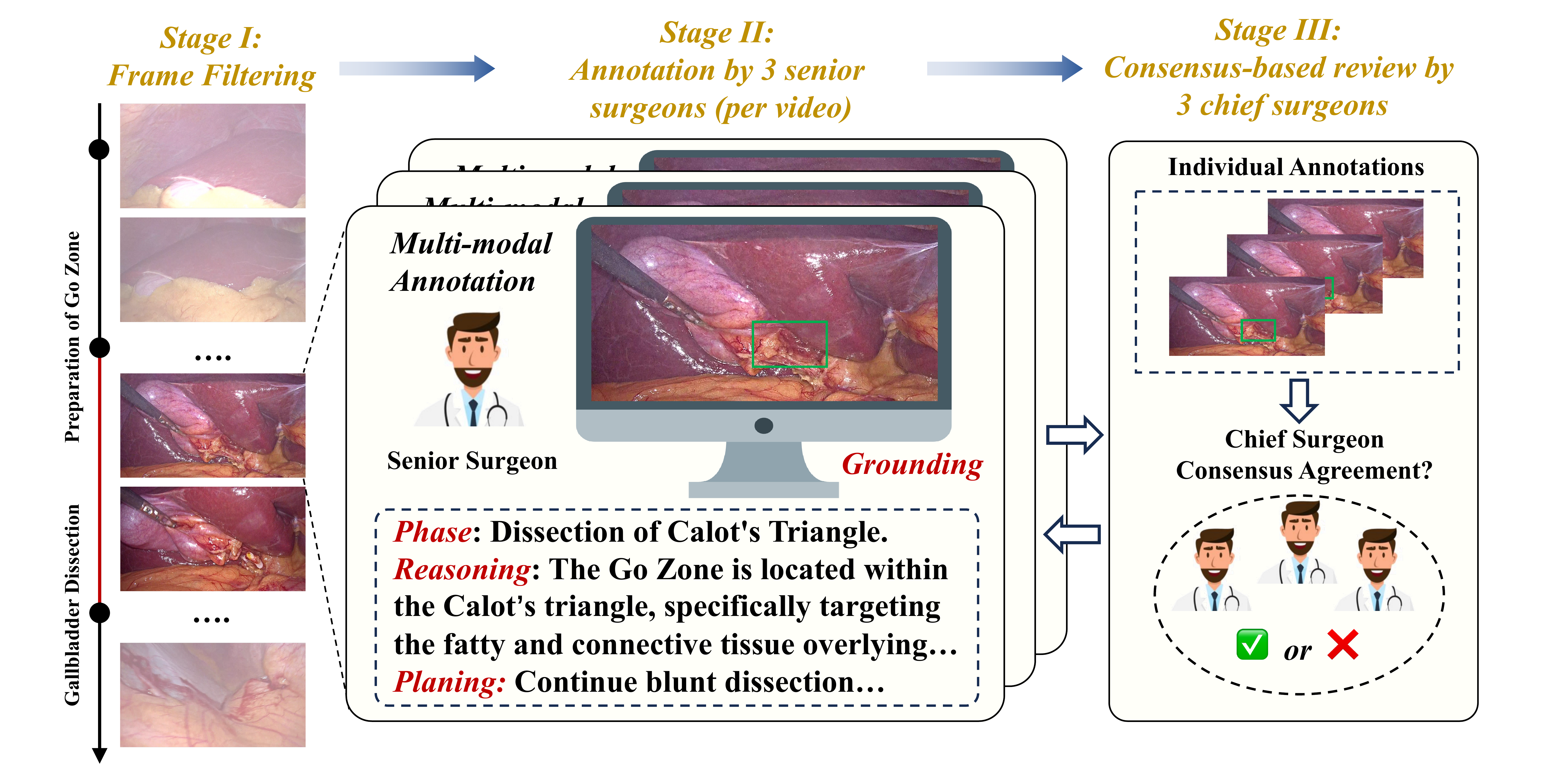}
    \caption{Illustration of the annotation and review pipeline}
    \label{fig:anpipe}
\end{figure}
\begin{enumerate}[label=\textbf{\textit{\alph*)}}]
\item \textbf{Surgical Phase Identification:} Selected frames were classified into specific procedural phases, \emph{i.e.}, Phase A - Preparation of the Go Zone, Phase B - Dissection of Calot’s Triangle, Phase C - Clip and Divide, and Phase D - Gallbladder Dissection from Liver Bed.
\item \textbf{Grounding of the Go Zone:} Bounding boxes were drawn around safe operation areas based on clinical experience adhere to the agreed definition. These annotations were executed using the \texttt{LabelMe} tool. Further, textual anatomic descriptions of the Go Zone are also included.
\item \textbf{Exposure Quality Reasoning:} Reasoning regarding the quality of Go Zone exposure was conducted based on visual cues from specific anatomical tissues.
\item \textbf{Planning - Action \& Risk Reminders:} Subsequent procedural steps were formulated at a fine-grained level based on temporal context and surgeon judgment. For instance, specific actions, such as applying traction to facilitate exposure, were annotated to reflect intraoperative decisions, accompanied by reminders of potential risks arising from incorrect maneuvers.
\end{enumerate}
For the textual components (Phase, Anatomic, Reasoning, and Planning), data were recorded via \texttt{Excel} and indexed by video frame number to ensure alignment with the visuals. A detailed summary is provided in Tab.~\ref{tab:dataset_stats}. The annotation process was conducted on the 2,686 previously selected frames, yielding corresponding total samples.

To guaranty data quality for benchmarking and maintain high consensus, a rigorous quality control architecture was adopted. Following the initial annotation, a mutual cross-check was first performed by the annotating surgeons for a format check. Subsequently, a comprehensive sample-by-sample review of the entire dataset was conducted by three chief surgeons, each possessing over 15 years of experience. Frames were included only upon unanimous agreement among reviewing surgeons.

After the review, 90 frame samples of 4 videos were identified for a refinement process. These adjustments were necessitated by discrepancies including, bounding boxes slightly exceeding anatomical boundaries and reasoning annotations that were deemed fragmented, requiring a modification into a  structured format. Following refinement, full consensus was achieved among all senior and chief surgeons without the need for further review, thereby ensuring the accuracy and adherence to surgical protocols.

\subsection{Analysis and Splits of Annotated Samples} In this section, the dataset is analyzed based on the four annotated phases (A through D), with detailed distributions presented in Fig. \ref{fig:all_stats}. It is observed that the preparation phase exhibits the shortest duration. Given that the exposure of the Go Zone in cholecystectomy is considered procedurally straightforward compared to other complex surgeries, this relative brevity is expected and is found to be aligned with routine clinical practice.

Valuable insights are further derived from the transition matrix. While the preparation phase is consistently followed by the dissection of Calot's triangle, an alternating pattern is frequently observed between the dissection of Calot's triangle and the clip and divide phase. This suggests that the procedural workflow is modulated by surgical complexity and the surgeon's execution strategy. While a linear progression may be achieved in standard cases or by highly experienced practitioners, this iterative alternation reflects the adaptive measures required when increased difficulty is encountered. This variability highlights the necessity of the proposed \textit{Reasoning Go}, which is intended to assist surgeons by reducing intraoperative cognitive burden and facilitating a more fluid surgical workflow or provide education.

Finally, it is observed that, in terms of duration, the gallbladder dissection phase records the longest average duration. Conversely, significant variance is noted in the dissection of Calot's triangle, where the presence of lower extreme values suggests that rapid phase transitions exist in scenarios.

\paragraph{Train and Test Splits} 
The dataset was partitioned into training and testing splits at the video level to ensure robust evaluation on unseen patients. Specifically, 4 of the 21 videos were allocated to the testing set, comprising a total of 495 samples. The remaining 17 videos comprising 2,191 frames were designated for the training set.

\subsection{Benchmark Problem Formulation}
Instead of modeling Go Zone reasoning as a static grounding task, it is formulated as a \textbf{conditioned derivative} of the evolving surgical context. Precise phase identification serves as a contextual anchor rather than a final output, facilitating phase-conditioned reasoning to deliver effective intra-operative guidance, since, crucially, a failure in phase recognition renders any downstream grounding or reasoning \textbf{futile}, since a Go Zone predicted under an incorrect phase assumption is clinically meaningless. 

To enforce this dependency and align with the cognitive workflow validated by surgeons, the benchmark is structured not as a one-shot reasoning task, but as a sequential decision process unfolding as \textbf{phase-then-go}. This design necessitates that the phase-level understanding derived from the initial inference is explicitly incorporated into the subsequent Go Zone reasoning as:
\begin{equation}
    P(\mathbf{b}, p | \mathcal{I}) = \underbrace{P(p | \mathcal{I})}_{\text{Phase Recognition}} \cdot \underbrace{P(\mathbf{b} | \mathcal{I}, p, \mathcal{D}(p))}_{\text{Contextual Reasoning}},
\end{equation}
where $\mathbf{b}$ denotes the Go Zone reasoning and $\mathcal{I}$ represents the input frame. $\mathcal{D}(p)$ signifies the Go Zone definition and operative criteria corresponding to phase $p$. This formulation transforms the problem from implicit pattern matching into a structured inference task, ensuring that the Go Zone is strictly derived from the distinct anatomical rules of the identified phase.

\subsection{Evaluation Protocols}
A hierarchical evaluation strategy is adopted to reflect the sequential nature of the benchmark. While standard grounding metrics, such as accuracy at IoU thresholds (Acc@$_{0.25}$, mA@$_{0.25:0.5}$), center distance normalized by image diagonal ($\Delta_{\text{cen}}$), and mean IoU, are reported, they are deemed insufficient where valid grounding is contingent upon correct phase identification. To address this, \textbf{\textit{conditioned metrics}} are introduced, where spatial reasoning is evaluated solely on samples with correctly predicted phase, to simulate the scenarios when phase metadata is provided for educational purposes. Furthermore, \textbf{\textit{hardcore metrics}} (HA$_{0.25}$, HmIoU) are proposed to measure end-to-end reliability. In this framework, grounding outputs associated with erroneous phase identification are penalized as invalid regardless of spatial accuracy, thereby capturing the joint success of phase recognition and Go Zone. For the explicit assessment of reasoning capabilities, a small-scale, double-blind review by surgeons is recommended, wherein generated outputs are scored based on clinical correctness.

\section{\textit{SurGo-R1} Baseline Methodology} 
\subsection{Overview of \textit{SurGo-R1}} 
\begin{figure}[!t]
    \centering
    \includegraphics[width=\linewidth]{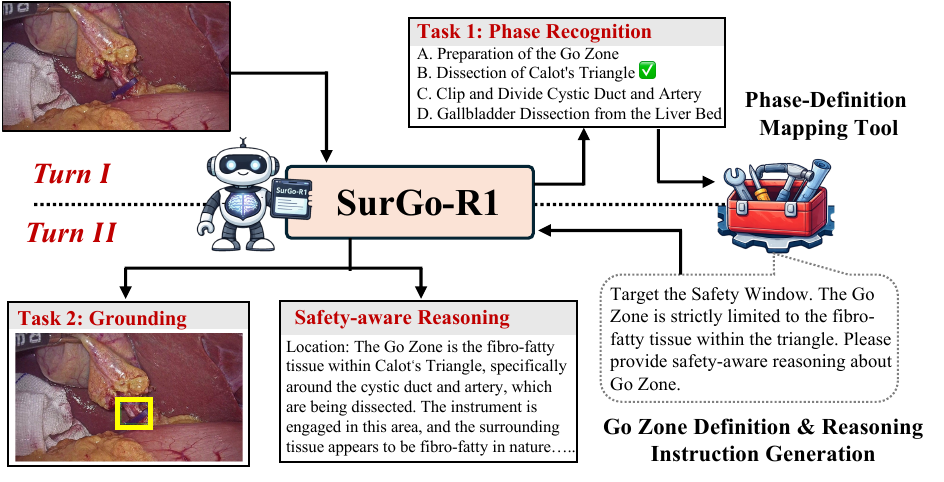}
    \caption{Overview of SurGo-R1 and phase-then-go pipeline.}
    \label{fig:net}
\end{figure}

To practically instantiate the proposed phase-then-go logic, \textit{SurGo-R1}, a multi-modal reasoning framework optimized via GRPO, is introduced as a baseline to serve as a reference standard for the community and to facilitate future methodological advancements. In Turn 1, the model answers a phase Multiple Choice Question (MCQ), compelling the model to prioritize procedural context as a prerequisite before reasoning. In Turn 2 then performs reasoning and grounding conditioned on the predicted phase with tool-provided definitions, as in Fig.~\ref{fig:net}

\subsubsection{Prerequisite Turn: Contextual Priming}
In the first turn, we establish the procedural context by presenting the model with a surgical scene frame and candidate phase labels. Framed as an MCQ, the model identifies the current phase based on visual cues. To incentivise precise phase recognition, we train with GRPO using a strict binary accuracy reward:
$\mathbf{R}_{acc} = \begin{cases}
1 & \text{if } \hat{p} = p \\
0 & \text{otherwise}
\end{cases}$
, where the reward is 1 if the predicted phase $\hat{p}$ matches the ground truth $p$, and 0 otherwise. Having model explicitly training with phase MCQ ensures the model prioritizing surgical phase recognition as a foundation.

\subsubsection{Reasoning Turn: Surgical Analysis}
Given the predicted phase $\hat{p}$, the model proceeds to Go Zone reasoning and grounding. Following an agentic workflow, it invokes a phase-definition mapping tool that retrieves procedure-specific constraints $\mathcal{D}_{\hat{p}}$, providing structured context for the visual input. Guided by this definition, the model generates a structured output $\mathbf{b}$, comprising a reasoning rationale across four dimensions related to surgical safety, as in  (1) \textbf{Location} relative to anatomical landmarks, (2) \textbf{Exposure} adequacy, (3) \textbf{Next action}, and (4) \textbf{Critical risk}, alongside the final Go Zone coordinate prediction, grounding the spatial output in a structured surgical assessment.

\paragraph{Phase-Definition Mapping Tool}
The agentic workflow is implemented through a dynamic contextual prompting mechanism that functions as an intermediate supervisor, with distinct behaviors adopted for training and inference. During the training stage, the predicted phase $\hat{p}$ is compared against the ground truth $y$. To ensure optimal conditioning for the subsequent reasoning task, any classification errors are rectified, and the definition $\mathcal{D}_y$ corresponding to the correct phase is injected into the prompt. Conversely, during the inference stage, the process operates autonomously: the definition $\mathcal{D}_{\hat{p}}$ is dynamically retrieved and inserted based solely on the model's prediction, regardless of its accuracy. Content of this tool is provided in the appendix.

This differential strategy provides two complementary benefits: \textit{(a)} \textbf{Intensified Spatial Alignment:} By rectifying phase errors during training, it is ensured that the reasoning module is consistently exposed to valid procedural contexts. This isolates spatial learning from classification noise, thereby enabling the model to master the challenging text-to-image grounding task without interference from upstream errors. \textit{(b)} \textbf{Adaptive Context Integration:} This mechanism trains the model to strictly condition its spatial reasoning on the provided textual definition. Consequently, during inference, the model is capable of generating grounding outputs that are logically consistent with its own phase predictions.

\paragraph{Reward Modeling}
In the reasoning turn, we continue to employ GRPO to optimize the generation of the clinical rationale, covering the above-mentioned four dimensions, and Go Zone grounding. We design a composite reward function to provide supervision signals while balancing computational efficiency.

\begin{table*}[!t]
    \centering
    \caption{Quantitative results on the \textit{ResGo} dataset. The best results are highlighted in \textbf{bold}.}
    \label{tab:reasoning_go_results}
    \resizebox{\textwidth}{!}{
    \begin{tabular}{l | c | c c c c | c c c c | c c }
        \toprule
        \multirow{2}{*}{\textbf{Method}} & \textbf{Phase} & \multicolumn{4}{c|}{\textbf{Grounding}} & \multicolumn{4}{c|}{\textbf{Conditioned}} & \multicolumn{2}{c}{\textbf{Hardcore}} \\
        \cmidrule(lr){2-2} \cmidrule(lr){3-6} \cmidrule(lr){7-10} \cmidrule(lr){11-12}
         & Acc & Acc@$_{0.25}$ & mA@$_{0.25:0.5}$ & $\Delta_{cen}\downarrow$ & mIoU & CA$_{0.25}$ & CA$_{0.25:0.5}$ & C$\Delta_{cen}\downarrow$ & CmIoU & HA$_{0.25}$ & HmIoU \\
        \midrule
        \multicolumn{7}{l}{\textit{Generalist VLMs}} \\
        Qwen3-VL-4B-Ins & 36.5 & 14.7 & 7.61 & 11.7 & 12.3 & 21.5 & 11.5 & 8.84 & 15.8 & 7.88 & 5.79 \\
        Qwen3-VL-8B-Ins & 39.3 & 16.9 & 7.83 & 10.9 & 13.2 & 14.8 & 6.24 & 9.83 & 12.5 & 5.86 & 4.94 \\
        Qwen3-VL-32B-Ins & 53.1 & 16.3 & 8.01 & 9.29 & 14.2 & 15.5 & 7.67 & 9.26 & 13.5 & 8.28 & 7.20 \\
        Qwen3-VL-30B-A3B-Ins & 51.9 & 11.9 & 5.35 & 12.6 & 10.7 & 12.8 & 5.51 & 11.7 & 11.4 & 6.67 & 5.95 \\
        Qwen3-VL-4B-Thi & 31.5 & 3.84 & 2.05 & 20.8 & 3.71 & 6.41 & 2.88 & 20.2 & 5.46 & 2.02 & 1.72 \\
        Qwen3-VL-8B-Thi & 34.1 & 12.3 & 5.79 & 16.7 & 8.04 & 12.4 & 6.21 & 18.2 & 7.18 & 4.24 & 2.45 \\
        InternVL3.5-8B & 33.5 & 7.27 & 3.37 & 17.0 & 5.81 & 6.63 & 3.11 & 15.9 & 6.21 & 2.22 & 2.08 \\
        InternVL3.5-14B & 35.3 & 4.85 & 2.12 & 16.2 & 5.46 & 5.14 & 1.62 & 14.7 & 6.33 & 1.82 & 2.24 \\
        Ovis2.5-2B & 22.6 & 1.01 & 0.34 & 24.1 & 1.29 & 0.97 & 0.33 & 22.6 & 1.04 & 0.27 & 0.39 \\
        Ovis2.5-9B & 42.8 & 1.82 & 0.57 & 15.1 & 2.08 & 1.42 & 0.47 & 14.4 & 2.05 & 0.61 & 0.88 \\
        \midrule
        \multicolumn{7}{l}{\textit{Specialist Models}} \\
        EndoChat & 30.5 & 0.81 & 0.17 & 21.0 & 2.23 & 1.32 & 0.33 & 20.7 & 2.46 & 0.40 & 0.75 \\
        Hulu-Med-7B & 37.5 & 4.04 & 1.55 & 10.5 & 8.03 & 5.38 & 2.06 & 10.3 & 8.02 & 2.02 & 3.01 \\
        Hulu-Med-14B & 31.3 & 0.81 & 0.24 & 10.9 & 5.05 & 0.65 & 0.11 & 11.2 & 4.68 & 0.20 & 1.46 \\
        QoQ-Med3-VL-8B & 37.7 & 15.9 & 5.93 & 10.2 & 12.1 & 13.7 & 5.44 & 9.25 & 12.4 & 5.05 & 4.69 \\
        \midrule
        \rowcolor[HTML]{ECF4FF} \textbf{\textit{SurGo-R1}} & \textbf{76.6} & \textbf{68.3} & \textbf{39.7} & \textbf{4.11} & \textbf{32.7} & \textbf{71.5} & \textbf{40.9} & \textbf{3.63} & \textbf{33.8} & \textbf{54.8} & \textbf{25.9} \\
        \bottomrule
    \end{tabular}}
\end{table*}

\textbf{Reasoning Reward:} To align rationales with clinical standards, we introduce a lightweight reward based on semantic entity matching. Instead of rigid traditional template comparisons, we employ scispaCy to dynamically extract core entities from the ground truth, including surgical targets, operative actions, and safety constraints. The reward computes recall of these entities in the generated output:
$\mathbf{R}_{\text{reason}} = \frac{|{w \mid w \in \mathcal{W} \cap \mathcal{K}}|}{|\mathcal{K}|}$
, where $\mathcal{K}$ is the set of extracted entities and $\mathcal{W}$ the unique tokens in the prediction. This reward serves as a semantic anchor during GRPO training, incentivizing the model to integrate key clinical concepts into rationales, without the overhead of LLM-based reward models.

\textbf{Grounding Rewards:} To supervise Go Zone localization, we combine two reward functions. The primary objective is standard Intersection-over-Union (IoU) for boundary alignment:
$\mathbf{R}_{\text{IoU}} = \frac{|\mathbf{b}_{\text{pred}} \cap \mathbf{b}_{\text{gt}}|}{|\mathbf{b}_{\text{pred}} \cup \mathbf{b}_{\text{gt}}|}.$
However, relying solely on IoU presents a significant drawback in the context of surgery where early predictions may not overlap with the ground truth at all, yielding zero reward and vanishing gradients. To address this, we introduce an auxiliary Center Distance Reward that provides dense supervision by measuring proximity between predicted and ground-truth centers:
$\mathbf{R}_{\text{dist}} = \exp\left(-\frac{| \mathbf{c}_{\text{pred}} - \mathbf{c}_{\text{gt}} |_2}{\tau}\right)$
, where $\tau$ is a scaling factor adjusted by the grounding nature of VLMs. This term encourages the model to shift toward the correct anatomical region even without precise overlap, maintaining gradient flow and stabilizing GRPO training.

\subsection{Implementation Details}
\textbf{\textit{Qwen3-VL-8B-Instruct}} is utilized as the foundation VLM, and a two-stage training pipeline tailored to the sequential task structure is followed.
\textbf{Stage 1: Phase Recognition.} The model is trained exclusively on the phase identification turn using GRPO with 4 generations, a temperature of 0.9, a learning rate of $2\times10^{-6}$, a batch size of 16, and 10 epochs. The reference penalty $\beta$ is set to 0.01. Rewards are comprised of $\mathbf{R}_{\text{acc}}$ and a format reward, with loss applied only to the first turn to enforce phase recognition.
\textbf{Stage 2: Multi-Turn Reasoning.} The full pipeline is then trained with rewards $\mathbf{R}_{\text{acc}}$, $\mathbf{R}_{\text{reason}}$, $\mathbf{R}_{\text{IoU}}$, $\mathbf{R}_{\text{dist}}$, and a format reward. $\tau=100$ is set to match Qwen3's scale of grounding convention. GRPO training is conducted using 4 generations and $\beta=0.01$ over 80 steps. Loss is applied across the full trajectory to reinforce the dependency between phase recognition and reasoning. Sequence-level importance sampling is applied throughout. All experiments are executed on 4 NVIDIA H200 GPUs, with 1 GPU allocated for vLLM-based rollout sampling and 3 GPUs utilized for updates.

\begin{figure*}[!t]
    \centering
    \includegraphics[width=\linewidth]{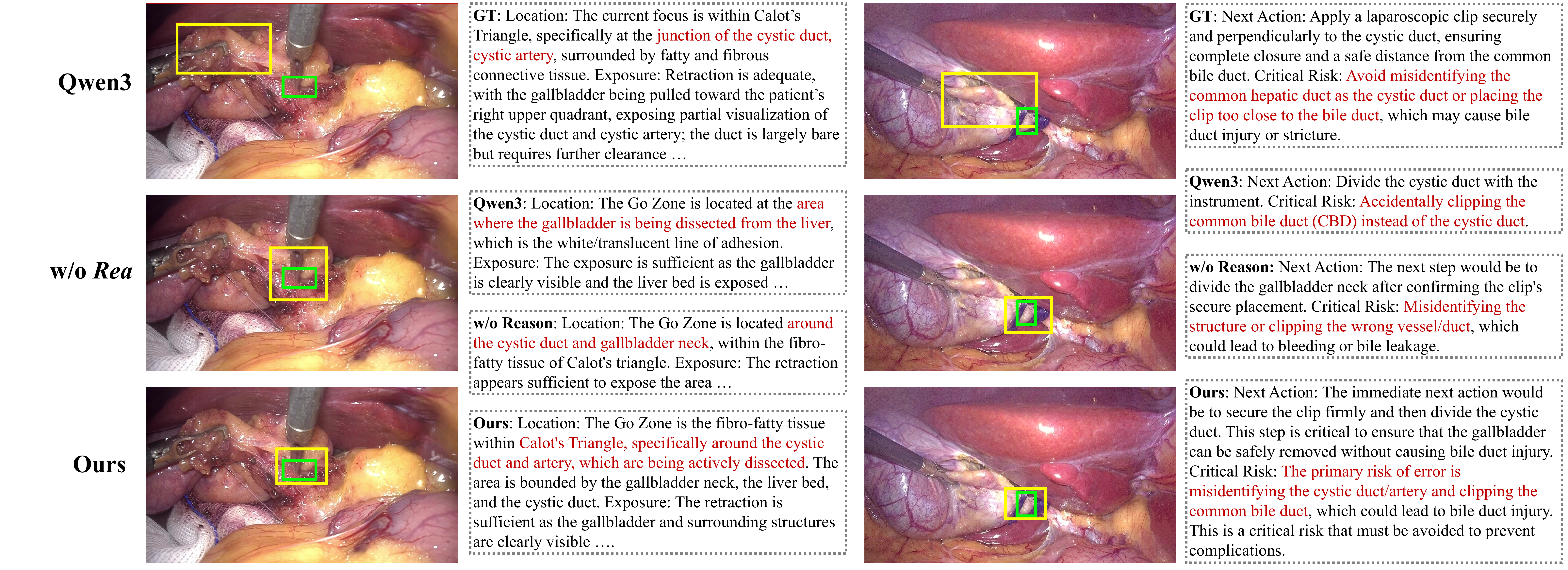}
    \caption{Qualitative results of reasoning and grounding. Models trained without $\mathbf{R}_{\text{reason}}$ are denoted as w/o Rea or w/o Reason. Yellow and green bounding boxes represent model predictions and ground truths, respectively.}
    \label{fig:vis}
\end{figure*}

\section{Experiments}

\subsection{Overall Performance Comparison}
To evaluate the capabilities of contemporary VLMs \cite{bai2025qwen3vl, wang2025internvl3,lu2025ovis2,wang2025endochat,jiang2025hulu,dai2025qoq} on the proposed benchmark, comprehensive inference experiments were conducted. It is noted that the evaluation protocol is strictly aligned with the hierarchical \textit{phase-then-go} formulation. This includes the utilization of the definition-tool, ensuring that the reasoning process during inference is structurally consistent with the proposed framework.

As evidenced in Tab.~\ref{tab:reasoning_go_results}, significant deficiencies are observed in existing generalist and specialist models regarding their capacity to align procedural phase knowledge with Go Zone grounding and safety reasoning. These limitations, particularly in ensuring consistency, \textbf{underscore the critical value and need of the proposed \textit{ResGo} dataset for these gaps}. Conversely, the substantial performance gains achieved by \textit{SurGo-R1} validate the feasibility of this challenging task, thereby establishing a foundation to incentivize research.

\subsection{Qualitative Analysis and Reasoning Study}
Visualizations and qualitative results SurGo-R1 generated are in Fig.~\ref{fig:vis}. Noted that SurGo-R1 generates better reasoning answers with more comprehensive descriptions of locations and risk explanations, while w/o Rea tends to provide less informative and shorter answers.

To further evaluate the impact of $\mathbf{R}_{\text{reason}}$, reasoning outputs on the whole test set were collected and de-identified (A/B/C) for a blind review by three clinicians to review select rate, while accuracy rating was noted based on factual correctness (0 for errors/hallucinations, 1 otherwise). As observed in Tab.~\ref{tab:reasoning_ablation}, \textit{SurGo-R1} is preferred more often and achieves higher ratings, confirming the reasoning reward.

\begin{table}[!t]
    \centering
    \caption{\textbf{Ablations on the \textit{ResGo} dataset.} \checkmark indicates inclusion in the training process, while \ding{55} denotes exclusion.}
    \label{tab:ablation_study}
    \resizebox{\linewidth}{!}{
    \begin{tabular}{c c | c c c | c c} 
        \toprule
        \multirow{2}{*}{\textbf{$\mathbf{R}_{\text{dist}}$}} & \multirow{2}{*}{\textbf{Def-Tool}} & \multicolumn{3}{c|}{\textbf{Grounding}} & \multicolumn{2}{c}{\textbf{Hardcore}} \\
        \cmidrule(lr){3-5} \cmidrule(lr){6-7} 
        & & Acc@$_{0.25}$ & $\Delta_{cen}\downarrow$ & mIoU & HA$_{0.25}$ & HmIoU \\
        \midrule
        \ding{55} & \ding{55} & 56.3 & 5.77 & 23.2 & 43.5 & 16.9 \\
        \checkmark & \ding{55} & 58.4 & 5.23 & 23.6 & 44.2 & 17.5 \\
        \ding{55} & \checkmark & 67.8 & 4.35 & 32.1 & 53.1 & 25.0 \\
        \checkmark & w/o-rectify & 64.7 & 4.58 & 30.3 & 48.7 & 21.4 \\
        \midrule
        \rowcolor[HTML]{ECF4FF} \checkmark & \checkmark & \textbf{68.3} & \textbf{4.11} & \textbf{32.7} & \textbf{54.8} & \textbf{25.9} \\
        \bottomrule
    \end{tabular}
    }
\end{table}

\begin{table}[!t]
    \centering
    \caption{Comparison between Single- and Multi-turn performance.}
    \label{tab:turn_comparison}
    \resizebox{\linewidth}{!}{
    \begin{tabular}{l | c | c c c | c c} 
        \toprule
        \multirow{2}{*}{\textbf{Method}} & \textbf{Phase} & \multicolumn{3}{c|}{\textbf{Grounding}} & \multicolumn{2}{c}{\textbf{Hardcore}} \\
        \cmidrule(lr){2-2} \cmidrule(lr){3-5} \cmidrule(lr){6-7} 
        & Acc & Acc@$_{0.25}$ & $\Delta_{cen}\downarrow$ & mIoU & HA$_{0.25}$ & HmIoU \\
        \midrule
        Single-turn & 69.5 & 66.9 & 4.24 & 32.1 & 49.6 & 22.7 \\
        \midrule
        \rowcolor[HTML]{ECF4FF} Multi-turn (Ours) & \textbf{76.6} & \textbf{68.3} & \textbf{4.11} & \textbf{32.7} & \textbf{54.8} & \textbf{25.9} \\
        \bottomrule
    \end{tabular}
    }
\end{table}

\begin{table}[!t]
    \centering
    \caption{Ablation study on the reasoning reward ($\mathbf{R}_{\text{reason}}$).}
    \label{tab:reasoning_ablation}
    \resizebox{\linewidth}{!}{
    \begin{tabular}{l | c c } 
        \toprule
        \textbf{Setting} & \textbf{Selected Ratio (\%)} $\uparrow$ & \textbf{Acc. Rating} $\uparrow$ \\
        \midrule
        Qwen3-VL-8B-Ins & 0.27 & 10.7 \\
        w/o $\mathbf{R}_{\text{reason}}$  & 17.3 & 47.2 \\
        \midrule
        \rowcolor[HTML]{ECF4FF} \textbf{w/ $\mathbf{R}_{\text{reason}}$ (Ours)} & \textbf{79.9} & \textbf{52.5} \\
        \bottomrule
    \end{tabular}
    }
\end{table}

\subsection{Ablation Study}
To investigate the effectiveness of the proposed components, ablation studies were performed on the \textit{ResGo} dataset, with results summarized in Tab. \ref{tab:ablation_study}. It is observed that the introduction of the distance reward ($\mathbf{R}_{\text{dist}}$) yielded improvements in spatial precision. A more substantial gain was observed with the integration of the Phase-Definition Mapping Tool. By explicitly retrieving anatomical definitions based on phases, the semantic gap between visual features and medical knowledge was effectively bridged. Furthermore, the analysis highlights the critical role of the rectification mechanism since its exclusion during training was found to degrade performance by subjecting the grounding module to erroneous procedural contexts. Consequently, the optimal performance was achieved when both components were employed, validating the proposed design.

Furthermore, the efficacy of the proposed multi-turn framework was evaluated against a single-turn baseline, as detailed in Tab. \ref{tab:turn_comparison}. To ensure a fair comparison, both models were initialized from the checkpoint following the Stage 1 MCQ training. It is evident that the multi-turn approach outperforms the single-turn counterpart across all metrics, given that the single-turn model attempts to simultaneously resolve phase identification and spatial grounding, which hinders model from learning.

\section{Conclusion}
To address the demand for explainable guidance in MIS, \textit{ResGo} is introduced as the first multimodal cholecystectomy dataset pairing spatial Go Zone localization with clinical rationales in this paper. Within this framework, surgical safety and grounding are reformulated as a contextual reasoning task, where perception is integrated with phase-dependent annotations of risks and procedures. Building on this, \textit{SurGo-R1} is proposed, a reasoning-based VLM optimized via GRPO to generate interpretable guidance and ground Go Zones. By transitioning to a multi-turn reasoning pattern, the gap between visual perception and complex surgical decision-making is effectively bridged, establishing a new foundation for advancing surgical intelligence.

\section*{Impact Statement}
This work advances the field of surgical AI, with primary benefits targeting clinical decision support. We acknowledge that rigorous validation is a prerequisite for further development and deployment. Regarding societal impact, this specific study does not introduce unique ones necessitating distinct discussion.

\bibliography{example_paper}
\bibliographystyle{icml2026}

\newpage
\appendix
\onecolumn
\section*{Appendix}
\subsection*{Phase-Definition Mapping Tool}
The content utilized by the Phase-Definition Mapping Tool is derived from a curated lexicon of phase-specific definitions, as detailed in Table \ref{tab:supp_definitions}. Comprising distinct anatomical targets and safety-driven exclusion criteria, these descriptions are dynamically inserted into the \texttt{<Surgical Context>} in the prompt to explicitly condition the model's spatial grounding.

\begin{table}[!ht]
\centering
\caption{\textbf{Anatomical Go Zone Definitions for Phase-Guided Reasoning.} The content utilized by the Phase-Definition Mapping Tool is presented in this table, supplying specific operative cues due to the deficiency in surgical domain knowledge common to current VLMs.}
\label{tab:supp_definitions}
\resizebox{\textwidth}{!}{%
\begin{tabular}{l p{11.5cm}}
\toprule
\textbf{Procedural Phase} & \textbf{Go Zone Definition \& Reasoning Instruction} \\
\midrule
\textbf{Preparation of the Go Zone} &
Target the Calot's triangle area. The Go Zone is defined as the peritoneum overlying the cystic duct and artery junction, where the initial dissection must commence to expose the underlying structures. \\
\midrule
\textbf{Dissection of Calot's Triangle} &
Target the ``Safety Window.'' The Go Zone is strictly limited to the fibro-fatty tissue within the triangle. It explicitly excludes the liver bed (cystic plate) and the common bile duct (CBD) to prevent bile duct injury. \\
\midrule
\textbf{Clip and Divide} &
Target the skeletonized cystic duct and artery. The Go Zone is the specific segment of these structures that is free of surrounding tissue and distinct from the CBD, rendering it suitable for safe clip application. \\
\midrule
\textbf{Gallbladder Dissection} &
Target the connective tissue plane (areolar tissue). The Go Zone is identified as the white or translucent line of adhesion separating the gallbladder from the liver bed. \\
\bottomrule
\end{tabular}%
}
\end{table}
\subsection*{Reasoning Turn Prompt}
We use the following template for the phase-guided reasoning turn that conditions the model on the predicted procedural phase and its corresponding definition retrieved from Phase-Definition Mapping Tool. Curly-braced fields (\texttt{\{...\}}) are populated at inference time: \texttt{\{phase\_name\}} is the predicted phase name from Turn 1 output, \texttt{\{choice\_letter\}} is the multiple-choice identifier model selected, and \texttt{\{definition\}} is the phase-specific Go Zone definition and constraints provided by the Tool.

\begin{tcolorbox}[
  title=Reasoning Turn Prompt,
  colback=gray!3,
  colframe=gray!50,
  boxrule=0.6pt,
  arc=2mm,
  left=1.5mm,right=1.5mm,top=1mm,bottom=1mm,
  fonttitle=\bfseries,
  enhanced,
  breakable
]
\begin{lstlisting}[style=promptstyle]
### Surgical Context
**Identified Phase**: {phase_name} ({choice_letter})
**Definition**: {definition}

### Mission
Locate the "Go Zone" strictly based on the **Definition** above.

### Required Output Format
<thinking>
Briefly map the visual landmarks (e.g., liver edge, duct, instrument) to the text definition above.
</thinking>

<reasoning>
1. Location: (Anatomical location description of the Go Zone)
2. Exposure: (Is retraction sufficient?)
3. Next Action: (Immediate maneuver)
4. Critical Risk: (Primary risk of error)
</reasoning>

<answer>
[xmin, ymin, xmax, ymax]
</answer>
\end{lstlisting}
\end{tcolorbox}

\end{document}